\documentclass[11pt]{article}
\usepackage[hyperref]{ccl2024-en}
\usepackage{times}
\usepackage{url}
\usepackage{latexsym}
\usepackage{fancyhdr}
\usepackage{multirow}
\usepackage{graphicx}

\def\@fnsymbol#1{\ensuremath{\ifcase#1\or *\or \dagger\or \ddagger\or
   \mathsection\or \mathparagraph\or \|\or **\or \dagger\dagger
   \or \ddagger\ddagger \else\@ctrerr\fi}}
\newcommand{\ssymbol}[1]{^{\@fnsymbol{#1}}}

\title{Mitigating the Bias of Large Language Model Evaluation}

\author{
    Hongli Zhou$^1$, Hui Huang$^2$, Yunfei Long$^3$, Bing Xu$^2$, Conghui Zhu$^2$, \\
    {\bf Hailong Cao$^2$, Muyun Yang$^2$$\ssymbol{1}$, Tiejun Zhao$^2$} \\
    $^1$School of Architecture and Design, Harbin Institute of Technology, Harbin, China \\ 
    $^2$Faculty of Computing, Harbin Institute of Technology, Harbin, China \\
    $^3$University of Essex \\
    {\tt \{hongli.joe,huanghui\}@stu.hit.edu.cn;yl20051@essex.ac.uk;}\\
    {\tt \{hitxb,conghui,caohailong,yangmuyun,tjzhao\}@hit.edu.cn}
}
\begin{document}
\maketitle
\cclfootnote{$\ssymbol{1}${Corresponding author.}}
\setcounter{footnote}{0}
\begin{abstract}
  Recently, there has been a trend of evaluating the Large Language Model (LLM) quality in the flavor of LLM-as-a-Judge, namely leveraging another LLM to evaluate the current output quality. However, existing judges are proven to be biased, namely they would favor answers which present better superficial quality (such as verbosity, fluency) while ignoring the instruction following ability. In this work, we propose systematic research about the bias of LLM-as-a-Judge. Specifically, for closed-source judge models, we apply calibration to mitigate the significance of superficial quality, both on probability level and prompt level. For open-source judge models, we propose to mitigate the bias by contrastive training, with curated negative samples that deviate from instruction but present better superficial quality. We apply our methods on the bias evaluation benchmark, and experiment results show our methods mitigate the bias by a large margin while maintaining a satisfactory evaluation accuracy.
\end{abstract}

\section{Introduction}
\label{intro}

\cclfootnote{
    %
    %
    \hspace{-0.65cm}  
    \textcopyright 2024 China National Conference on Computational Linguistics

    \noindent Published under Creative Commons Attribution 4.0 International License
}

Recently, the evaluation for Large-scale Language Models (LLMs) has drawn considerable attention of research community \cite{liang2022holistic,chang2023survey}. As the capabilities of LLMs continue to develop across various tasks, it is essential to evaluate them from a comprehensive perspective \cite{qin2023chatgpt}. However, traditional evaluation metrics for generative models, such as BLEU \cite{papineni2002bleu} and ROUGE \cite{lin2004rouge}, only capture limited aspects of a model's performance.

Some research has proposed LLM-as-a-Judge \cite{alpaca_eval,zheng2023judging}, namely utilizing proprietary LLMs, especially GPT-4 \cite{achiam2023gpt}, to evaluate the LLM's response.  However, relying on external API for evaluation may introduce consideration about privacy leakage. Other works propose to fine-tune open-source models as external evaluators \cite{pandalm2024,zhu2023judgelm}. They construct evaluation data and fine-tune LLMs to endow them specialized evaluation ability comparable with GPT-4. However, the fine-tuned judges underperform GPT-4 in terms of fairness and scalability by a large margin \cite{huang2024empirical}.

As a pioneering research, LLMBar \cite{zeng2023llmbar} introduces the first meta-evaluation benchmark of LLM evaluation bias. The benchmark is adversarially designed to test whether LLM evaluators can detect instruction-following outputs. They apply their benchmark on existing LLM evaluators, and reveal that even state-of-the-art LLM evaluators struggle to provide unbiased evaluation on their benchmark. However, their study is mainly based on closed-source models, and they only provide several suggestions to mitigate the bias by prompt engineering.

In this work, we propose a systematic framework to mitigate the bias of LLM evaluators. For closed-source LLM judges, we propose to mitigate the bias by calibration. As the evaluation bias is generally presented as a higher score for superficial quality, we propose two ways to directly model the superficial quality, and subtract it from the final result. More specifically, for probability-based evaluators, we propose to model the superficial quality by utilizing the probability generated by the pre-trained model. For generation-based evaluators, we propose to model the superficial quality by probing the superficial attributes in the prompt template. For open-source LLM evaluators, we propose to mitigate the bias of LLM evaluators by pairwise contrastive training. Delicately curated negative samples are included in the training set where the answer deviates from the instruction but presents better superficial quality.

We evaluate our methods on the bias benchmark of LLMBar, which consists of one natural and four adversarial test sets to benchmark LLM evaluators' ability to detect instruction-following answers unbiasedly. Experiment results show that all of our proposed methods mitigate the evaluation bias on the adversarial sets by a large margin, while keeping a comparable evaluation performance on the natural set.

\section{Related Work}
Presently, evaluation research on LLMs can be grouped into three distinct methodologies: human evaluation, benchmark evaluation, and LLM-as-a-Judge.

The most straightforward approach to evaluating an LLM involves engaging dedicated evaluators or crowdsourcing to compare and assess the model's output \cite{guo2023close,bang-etal-2023-multitask,bubeck2023sparks}. However, human evaluation is labor-intensive and challenging to scale during the model development cycle. Additionally, human subjective factors can influence the evaluation results, making it difficult to guarantee the consistency and stability of model evaluations. Other scholars have devised standard test sets for the automatic evaluation of LLMs. These test sets typically encompass a range of tasks, assessing the LLM's performance by comparing its task outputs to standard answers \cite{liang2022holistic,yu2024kola,zhong2023agieval}. However, the benchmarks typically consist of problems of selection, which are not adapted to quantify the generation ability of LLMs.

Given the various issues with benchmark evaluation, some scholars have introduced the evaluation scheme with the name of LLM-as-a-Judge, which involves utilizing another proprietary LLMs to assess the output of the current model. For example, AlpacaEval \cite{alpaca_eval} constructs an evaluation set comprising 805 question-answer pairs and leverages GPT-4 to compare the winning rate of the model's answers against text-davinci-003. Similarly, MT-bench \cite{zheng2023judging} creates 80 high-quality multi-round test questions spanning eight common areas, including mathematics and reasoning. It then automatically grades the model's answers using GPT-4. 

However, relying on an external model to perform evaluation may introduce concerns about the privacy leakage. Also, the iterative versions of the API model also make the evaluation inconsistent. To tackle these challenges, some studies have advocated for the training of localized agent evaluation models. For instance, PandaLM \cite{pandalm2024} constructs data based on Alpaca instructions and GPT-3.5 annotation, and then fine-tunes LLaMA-7B \cite{pandalm2024} as a judge model. JudgeLM \cite{zhu2023judgelm} constructs data from GPT-4 annotations and fine-tunes a scalable judge model.

Despite the LLM-as-a-Judge becoming the default paradigm for LLM evaluation, various work has pointed out that the LLM-based judges are severely biased, namely the evaluator would make prediction based on spurious features (mainly the superficial quality) while ignoring the alignment with the instruction. Various biases are discussed in previous work, including position bias \cite{wang2023large}, verbosity bias \cite{saito2023verbosity}, formality bias and self-enhancement bias \cite{zheng2023judging}, etc. As a pioneering work, LLMBar proposes the first systematic evaluation of LLM evaluation bias. They created four adversarial test sets as the testbed for LLM's ability of making unbiased prediction. Their results indicate that LLM evaluators, even GPT-4, suffer severely from the over-reliance on the superficial quality. However, they only propose to compensate for the bias by prompt engineering, which can hardly be applied to fine-tuned judges. A more thorough bias mitigation framework is of urgent demand.

\section{Methodology}

\subsection{The Formal Definition of LLM-as-a-Judge}
\label{sec:formal_definition}

Given a pre-defined evaluation set which contains various instructions, the LLM can be evaluated in both pairwise and pointwise manners:

$$\mathrm{P}(A|I) = argmax\sum_{t=m}^{m+n}\log p(h_t|\mathrm{T}(I, A), \theta)$$

$$\mathrm{P}(A_1, A_2|I) = argmax\sum_{t=m}^{m+n}\log p(h_t|\mathrm{T}(I, A_1, A_2), \theta)$$

\noindent where $h_m$ to $h_{m+n}$ is the span of predicted tokens, $T(\cdot)$ denotes the prompt template which is designed according to the judge and the evaluation scheme, $I$ denotes the instruction, $A$ denotes the answer, and $\theta$ is the model parameters of LLM. When generating an evaluation through decoding, the token of the evaluation part will be generated by taking the argmax of the vocabulary probability of each step.

Moreover, denoted by \cite{fu2023gptscore}, the generated log-probabilities (if available), is also an informative quality indicator that can be leveraged for evaluation:

$$\mathrm{P}(A|I) = \sum_{t=m}^{m+n}w_t\log p(h_t|h_{<t}, \mathrm{T}(I, A), \theta)$$

\noindent where $h_m$ to $h_{m+n}$ is the span of target tokens, and $w_t$ is the weight of the token at position $t$.

As the LLM-as-a-Judge can rely on both general-purposed closed-source LLMs and specially fine-tuned open-source LLMs, we propose two different strategies dedicated to mitigate the LLM evaluation bias in different scenarios. In the next sections, we will introduce the two methods one by one.

\subsection{Online Mitigation by Calibration}

As the bias of LLM evaluation mainly comes from the over-reliance to the superficial quality, the bias mitigation can be achieved by calibration in an on-the-fly manner. Calibration is a widely adopted technique to curate the language model predictions \cite{zhao2021calibrate}. Neural autoregressive language models take a sequence of tokens as input and output a probability distribution over the next token. As LLMs are typically trained with maximizing the log-likelihood, its prediction tends to be biased towards tokens that are frequent in the training set. Therefore, the bias can be mitigated by firstly quantifying the bias lying in the prediction distribution, then make a explicit subtraction. More specifically, as discussed in Section \ref{sec:formal_definition}, there are two flavors of LLM-as-a-Judge, namely probability-based and generation-based, and we propose different quantification methods for the superficial quality, respectively.

For probability-based evaluation, we propose to quantify the superficial quality with pre-trained models. The reason behind it is simple, since LLM typically gains its instruction following ability during supervised fine-tuning (SFT), the difference between the prediction distribution of pre-trained and SFT model can be indicative of instruction alignment. If one instruction-answer pair is favored by the pre-trained model, but unfavored by the instruction-tuned model, we would deem the answer as holding fluency but no instruction alignment.

\begin{figure}[h]
    \centering
    \includegraphics[width=\textwidth]{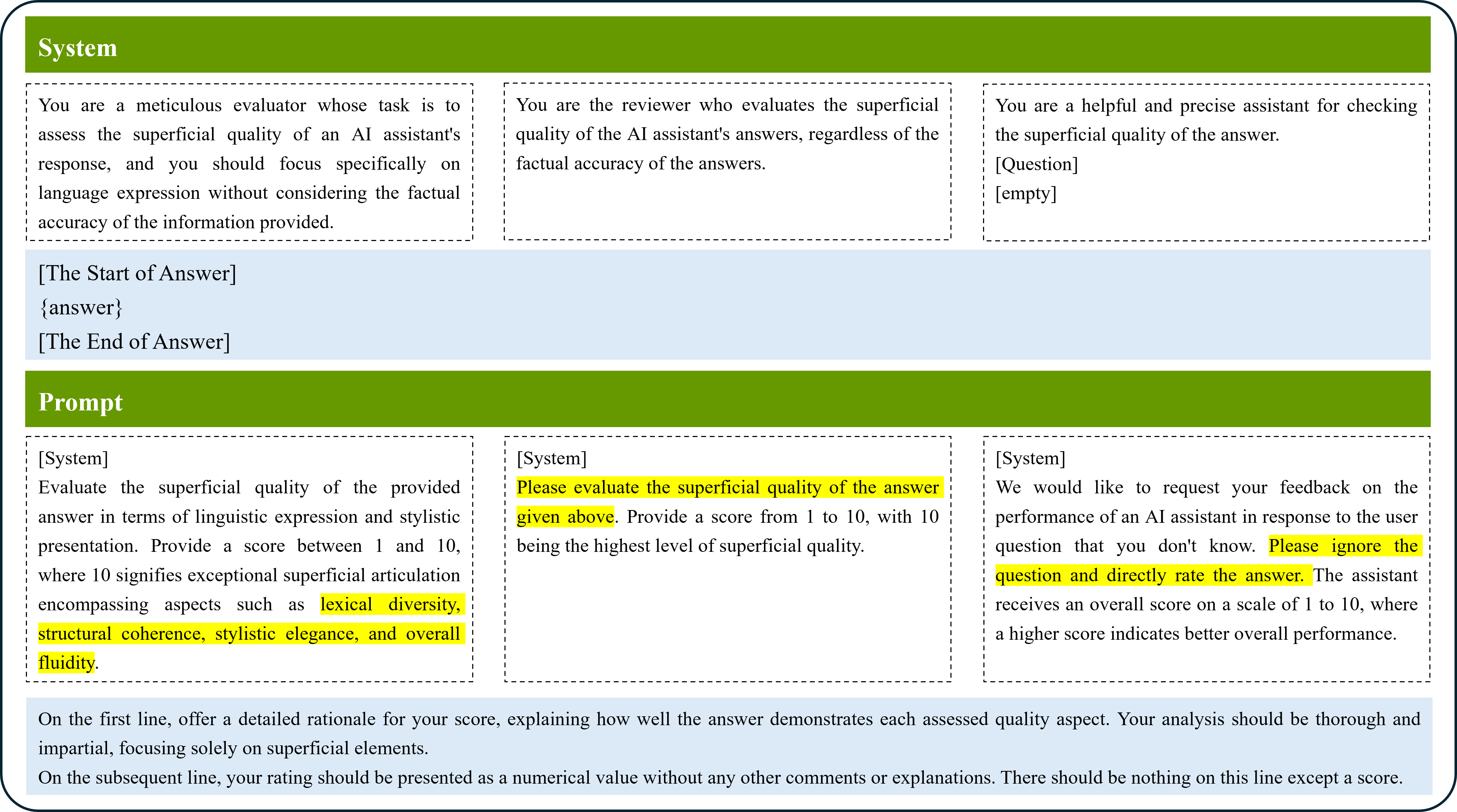}
    \caption{The three prompts for modeling superficial quality. The text in the blue blocks is the common part of the prompts, and together with the text in the dotted blocks, they constitute three types of prompts.}
    \label{fig:prompts}
\end{figure}

For generation-based evaluation, as the prediction logits is inaccessible, it is hard to calculate the distribution difference between pre-trained and SFT models. Therefore, we propose to directly model superficial quality by prompt engineering. More specifically, we design three prompts for quantifying the superficial quality, as shown in Figure \ref{fig:prompts}.

\begin{figure}[h]
    \centering
    \includegraphics[width=\textwidth]{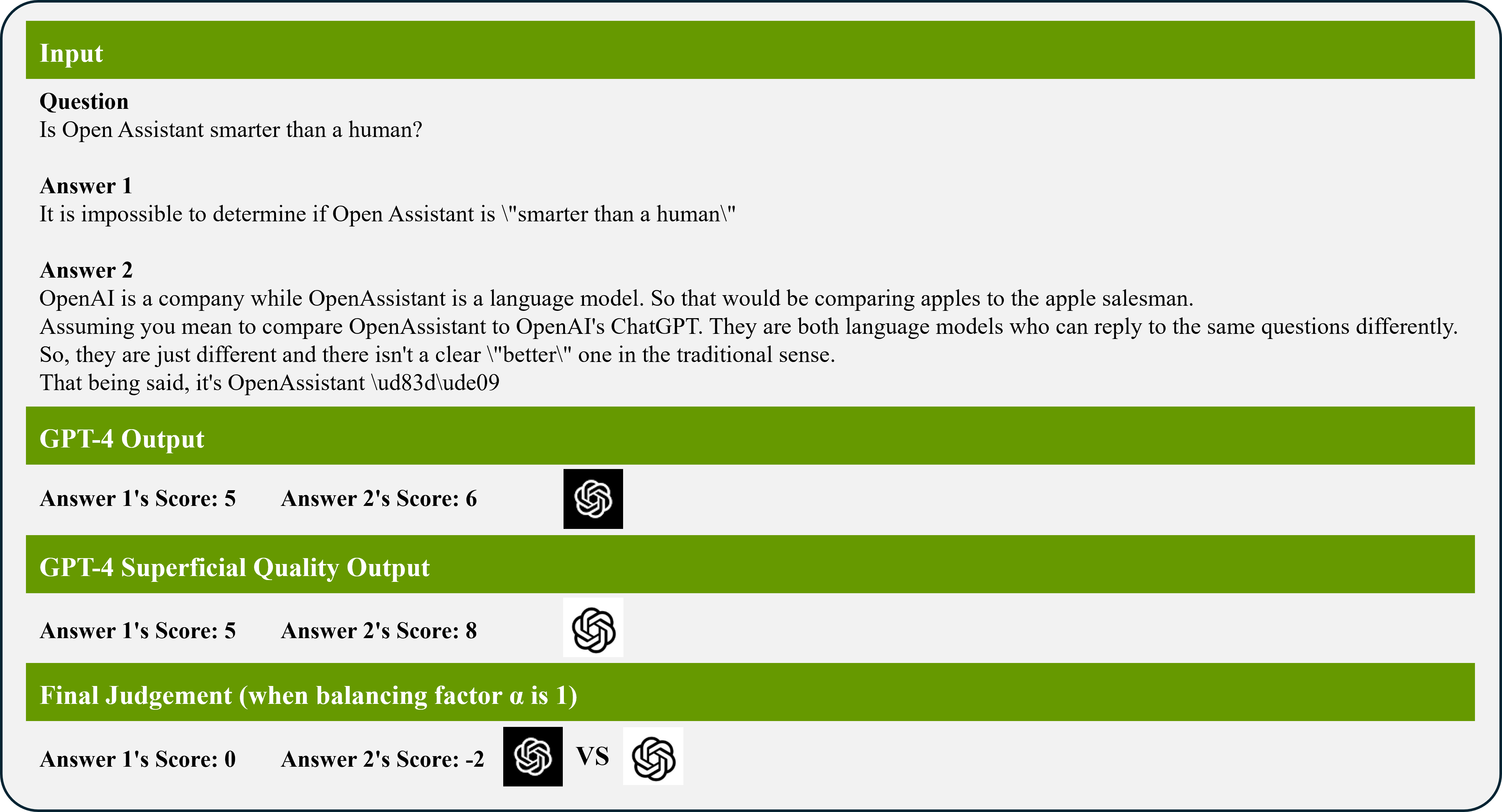}
    \caption{An example of online mitigation by calibration for generation-based evaluation. The correct output is answer 1 is better. It can be seen that the bias of LLM evaluation is effectively mitigated by subtracting the superficial quality.}
    \label{fig:superficial-quality}
\end{figure}

Finally, the attribute of superficial quality is excluded from the evaluation result by subtraction:

\begin{equation}
    \mathrm{\hat{P}(A|I)} = Normalize(\mathrm{P}(A|I)) - \alpha Normalize(\mathrm{Q}(A|I))
\label{equ:equation}
\end{equation}

$$Normalize(x_i) = \frac{Max(x_i)-x_i}{Max(x_i)-Min(x_i)}$$

\noindent where $Q(A|I)$ denotes the scores derived from superficial quality modeling, and $\alpha$ is a balancing factor, and $Normalize(\cdot)$ is to normalize predictions of different models into the same scale. Intuitively, if an instruction-answer pair holds high superficial quality (such as fluency, verbosity, engaging tones, etc), but misaligns with the instruction, it may receive a higher score in $P(A|I)$, but also a higher score in $Q(A|I)$, therefore our calibrated evaluation would be less biased, as illustrated in Figure \ref{fig:superficial-quality}.

\subsection{Offline Mitigation by Contrastive Training}

\begin{figure}[h]
    \centering
    \includegraphics[width=\textwidth]{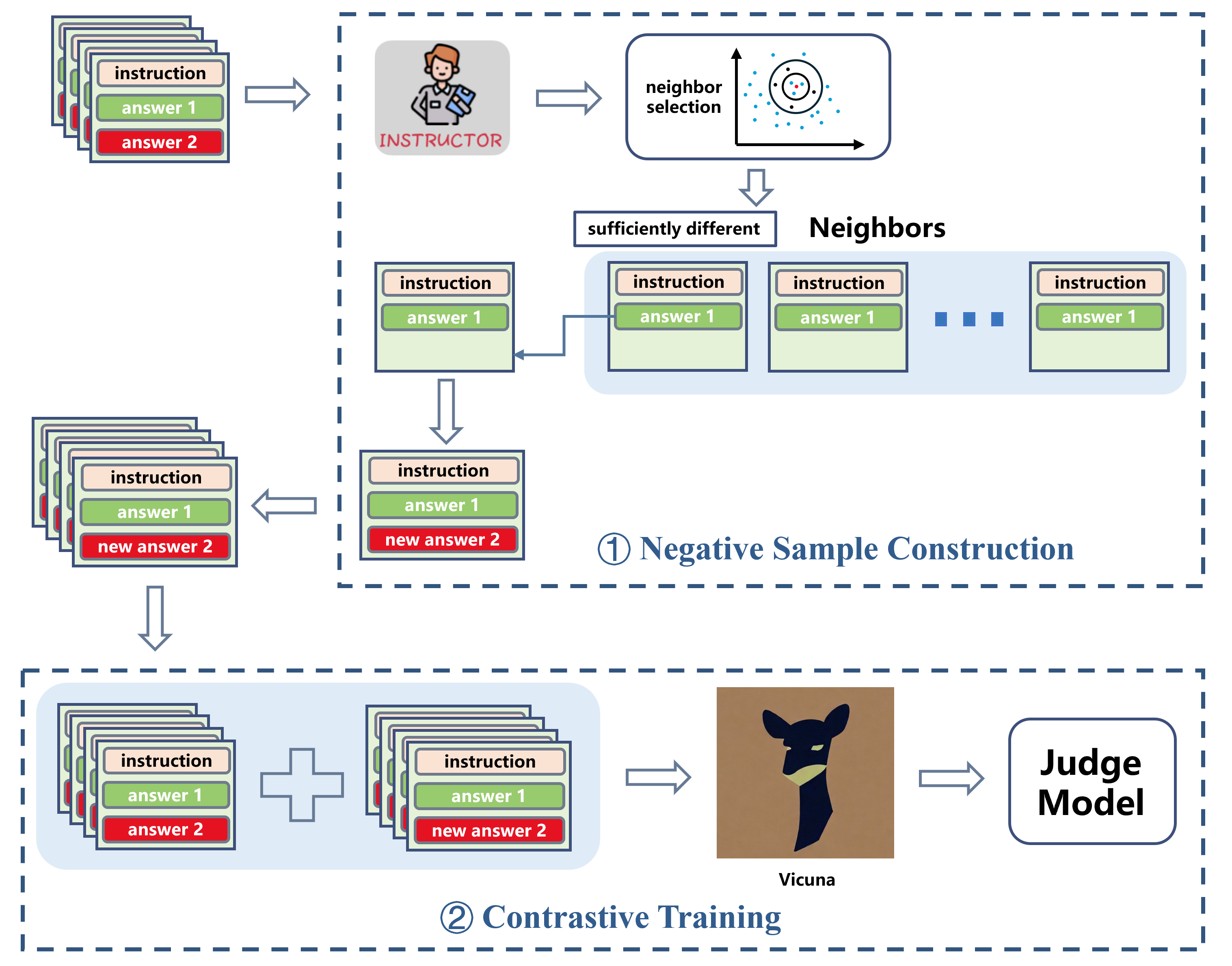}
    \caption{The pipeline of our offline mitigation. The first step is to construct negative samples in the original dataset D, and the second step is to perform contrastive training with the original data and the newly constructed data.}
    \label{fig:train}
\end{figure}

For fine-tuned LLM judges, both the checkpoint and the training data are openly-available. Therefore, it is possible to mitigate the bias in an offline manner, by preventing the model from over-relying on the spurious features for prediction. More specifically, we apply contrastive training on the LLM judge with delicately curated adversarial negative samples. Our method consists of the following two steps, namely negative sample construction and contrastive training, as shown in Figure \ref{fig:train}.

Firstly, we would like to create negative samples which are misaligned with the instruction, but hold better superficial quality. Given an instruction $I \in D$ where $D$ is its corresponding dataset, we retrieve a closely related yet sufficiently different instruction $I'$ from the same dataset D:

$$ I' = \mathrm{argmax}_{I'\in D, sim(I, I')<\epsilon}{sim(I, I')} $$
    
\noindent where $sim(\cdot, \cdot)$ is the cosine similarity measured by Instructor \cite{su-etal-2023-one}, a sentence embedding model specifically designed for modeling instruction intentions, and $\epsilon$ is a threshold to ensure that $I'$ and $I$ are semantically different enough. Then, in combination with the preferred answer $A$, this process enables us to create triplets $(I, A, A')$, with the expectation that $A$ faithfully follows the instruction $I$, while $A'$ deviates from $I$.

Secondly, the judge model is trained based on the combination of both contrastive samples and natural samples. We followed the instruction tuning paradigm, format all datasets to follow a chatbot-style schema to unify the varied styles and formats of the instruction datasets. During this process, the judge model can both see natural triplets $(I, A, A')$ with the $A$ holds better superficial quality and better instruction following, and also see adversarial triplets $(I, A, A')$ with the $A'$ is more fluent, verbose but deviates from the instruction intentions, therefore the judge model would learn an unbiased prediction pattern, avoiding making prediction solely based on superficial quality. Notice during this process we does not introduce extra annotation, only constructing negative samples based on the available data, therefore our method is both effective and practicable.

\section{Experiment}
\subsection{Set-up}

Our experiments are based on LLMBar \cite{zeng2023llmbar}, which consists a Natural set and four Adversarial sets. The Natural set collects and filters preference data from existing benchmarks, aiming to gauge evaluator performance in real-world distributions. Conversely, the Adversarial set comprises adversarially crafted instances that tend to confound less adept evaluators. There are four adversarial sets, namely Neighbor, GPTInst, GPTOut, Manual, which are created in different methods to probe the evaluator's bias from different perspectives.

For fine-tuning the open-source judges, we leverage the data from JudgeLM \cite{zhu2023judgelm}, which contains 100K judge samples, with the instructions sampled from a large-scale set that contains various instruction datasets, and the answers are generated from leading open-source LLMs, and the evaluations are annotated by GPT-4. Due to the limitation of computational resource, we sample 20K samples for negative sample construction.

For online mitigation by calibration, we conduct experiments on two widely used models, namely Text-davinci-003\footnote{https://platform.openai.com/docs/models/gpt-3-5} and GPT-4-1106-preview\footnote{https://platform.openai.com/docs/models/gpt-4-1106-preview}, both have gone through supervised fine-tuning and gained instruction following ability. For probability calibration, we use Davici-002 to quantify the superficial quality, which is the non-instruction tuned version of Text-davinci-003.

For offline mitigation by contrastive training, we follow the setting of JudgeLM \cite{zhu2023judgelm}, which is a specifically fine-tuned judge model based on Vicuna-7B \cite{vicuna2023}. We conduct negative sample generation inside the dataset, therefore no extra annotation is required.

\subsection{Main Results}

\begin{table}[]
\resizebox{1.0\textwidth}{!}{
\begin{tabular}{c|c|c|c|cccc|c}
\hline
\textbf{Model}             & \textbf{Method}             & \textbf{Mitigation} & \textbf{Natural} & \textbf{Neighbor} & \textbf{Manual} & \textbf{GPTOut} & \textbf{GPTInst} & \textbf{Average} \\ \hline
\multirow{2}{*}{T003}      & \multirow{2}{*}{logprobs}   & N/A                 & 70.00            & 43.33             & 43.47           & 48.94           & 38.67            & 48.88            \\
                           &                             & prob-calib          & \textbf{71.50}   & \textbf{72.83}    & \textbf{65.21}  & \textbf{57.20}  & \textbf{68.38}   & \textbf{67.02}   \\ \hline
\multirow{2}{*}{T003}      & \multirow{2}{*}{genlogits}  & N/A                 & \textbf{76.00}   & 31.68             & 28.57           & \textbf{51.35}  & 27.33            & 42.99            \\
                           &                             & prob-calib          & 72.50            & \textbf{69.27}    & \textbf{72.27}  & 48.94           & \textbf{71.35}   & \textbf{66.87}   \\ \hline
D002                       & logprobs                    & N/A                 & 64.50            & 37.31             & 37.35           & 36.30           & 28.79            & 40.85            \\ \hline
\multirow{4}{*}{GPT-4}     & \multirow{4}{*}{generation} & N/A                 & \textbf{88.00}   & 37.31             & 56.52           & 79.79           & 64.13            & 65.15            \\
                           &                             & prompt-calib1       & 83.50            & \textbf{45.52}    & \textbf{65.21}  & 84.04           & 70.65            & \textbf{69.78}   \\
                           &                             & prompt-calib2       & 83.00            & 45.52             & 64.13           & 84.04           & 69.02            & 69.14            \\
                           &                             & prompt-calib3       & 82.00            & 44.40             & 61.96           & \textbf{86.17}  & \textbf{73.91}   & 69.69            \\ \hline
\multirow{3}{*}{Vicuna-7B} & \multirow{3}{*}{generation} & vanilla             & \textbf{71.50}   & 25.00             & 29.35           & \textbf{48.94}  & 23.91            & 39.74            \\
                           &                             & contrast (inst)     & 70.00            & 61.57             & 35.87           & 42.55           & 64.13            & 54.82            \\
                           &                             & contrast (answer)   & 69.00            & \textbf{65.67}    & \textbf{39.13}  & 39.36           & \textbf{72.83}   & \textbf{57.20}   \\ \hline
\end{tabular}}
\caption{Experiment results of different bias mitigation methods on LLMBar}
\label{tab:main-result}
\end{table}

As can be seen in Table \ref{tab:main-result}, both of our proposed methods managed to achieve a higher performance on the adversarial sets,
while keeping a decent evaluation performance on the natural set. This verifies that we succeed to mitigate the bias of LLM evaluation. 

Between the two calibration methods, probability calibration achieves better performance, with consistent improvement on all of the four adversarial test sets, while leading no degradation to the original evaluation performance. Prompt calibration also achieves improvement on all adversarial test sets, but the improvement is comparably marginal, and the performance on the natural set is also slightly degraded. This indicates that the bias towards superficial quality mainly comes from the pre-training process, and can not be effectively detected solely based on the SFT model. However, it is worth noting that the accuracy of GPT-4 on the GPTOut set is largely improved, which demonstrates that the self-enhancement bias is mitigated. On the other hand, by calculating the probability distribution difference between the pre-trained and SFT models, we can precisely model the instruction alignment, thereby achieving superior performance on both natural and adversarial test sets.

For the contrastive training based results, we also achieve improvement on the adversarial test sets. The only exception is GPTOut, where the contrastive trained model achieves inferior performance. We think this might be due to the results constructed by GPTOut are generated by GPT-4, which presents a clear pattern that cannot be imitated by automatic negative sampling. On other words, this subset contains more formality bias, which deserves our future exploration.

It can also be notified that it is difficult to improve the evaluation accuracy based on bias mitigation, as the superficial quality is also an important classification feature. Answer with better detailedness, fluency and formality is also favored by the human preference, therefore excluding these aspects in evaluating LLM performance is actually unreasonable.

\subsection{Compromise between Bias and Accuracy}

\begin{figure}[h]
    \centering
    \includegraphics[width=\textwidth]{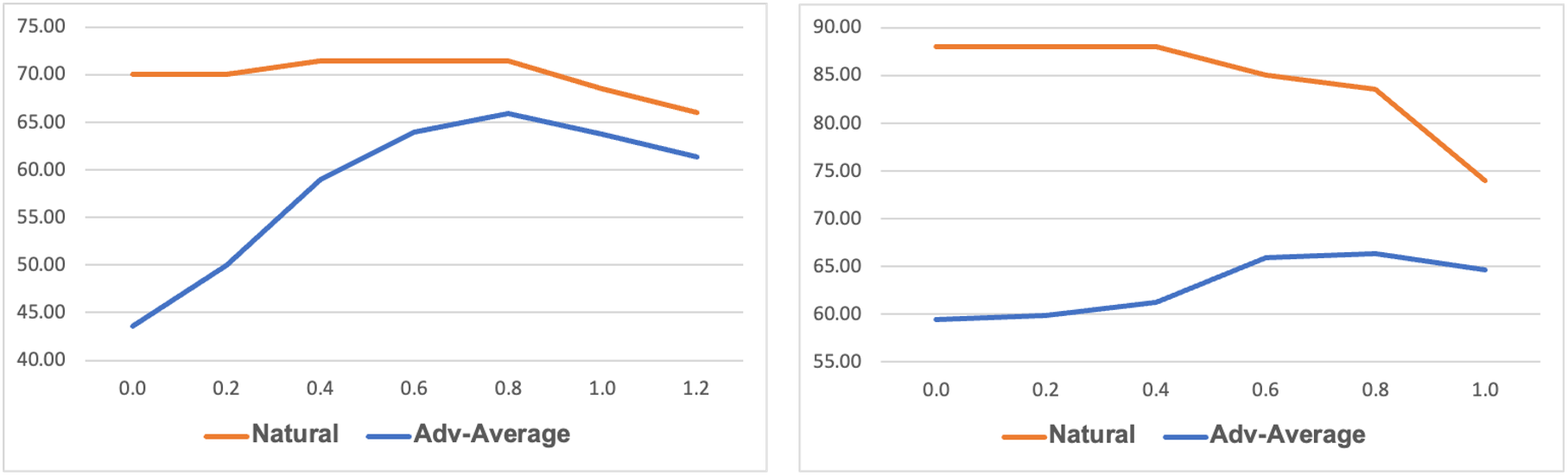}
    \caption{The variation of accuracy with respect to different coefficient. Left denotes the result of Text-davinci-003 on probability calibration, and right denotes the result of GPT-4 on prompt calibration.}
    \label{fig:trend}
\end{figure}

In this part, we would like to study the correlation between bias mitigation and evaluation accuracy. As we discussed before, while current LLM evaluators have been proved to be biased towards the superficial quality such as verbosity, formality, etc, these qualities are also important attributes of human preference and should not be excluded. We control the extent of bias mitigation by adjusting the coefficient factor $\alpha$ in Equation \ref{equ:equation}, and the change of accuracy with respect to $\alpha$ is shown in Figure \ref{fig:trend}.

As we can see, at the beginning of bias mitigation, the accuracy on the Natural set keeps stable, while the accuracy on the Adversarial set increase gradually. But after the factor is larger than 0.8, the accuracy on both  Natural set and Adversarial set would decrease. This verifies our hypothesis that there exists a compromise between bias mitigation and evaluation bias. Bias mitigation, to a certain extend, is helpful for improving the evaluation accuracy, but too much mitigation would lead to performance degradation eventually, 
and the superficial quality should not be excluded from the prediction process.

\subsection{Negative Sampling for Contrastive Training}

For contrastive training, the most important factor is the construction of negative samples. The negative samples should hold better superficial quality, while maintaining mismatch with the instruction. There is a compromise that lies here: The negative sample should not deviate from the instruction too much to improve the difficulty of contrastive training, neither should it be too close to the instruction to keep the negativity. Moreover, we also want to ensure the negative sample always holds better superficial quality. To achieve the most effective negative sampling, we conducted the following experiments:

\begin{enumerate}
    \item Adjacent Instruction: Select the most similar instruction within a scale;
    \item Adjacent Instruction + Adjacent Answer: Similar to 1, but this time directly select the answer that is most similar to the correct answer in the dataset within a scale, and add the data of the Adjacent Instruction;
    \item Adjacent Instruction + Superficial Scorer (Max): Train a classifier to discern the samples with better superficial quality. To achieve this goal, we train a binary classifier with only the target input, therefore the obtained classifier can only predict the evaluation based on the target input. After that, the classifier is used to select the sample with better superficial quality within the scale;
    \item Adjacent Instruction + Superficial Scorer (Avg): Similar to 3, but the score of superficial quality and semantic similarity is averaged for sampling;
\end{enumerate}

\begin{table}[]
\resizebox{1.0\textwidth}{!}{
\begin{tabular}{c|c|c|ccccc|c}
\hline
\textbf{Model}             & \textbf{Method}             & \textbf{Mitigation}  & \textbf{Natural} & \textbf{Neighbor} & \textbf{Manual} & \textbf{GPTOut} & \textbf{GPTInst} & \textbf{Average} \\ \hline
\multirow{5}{*}{Vicuna-7B} & \multirow{5}{*}{generation} & Vanilla              & \textbf{71.50}   & 25.00             & 29.35           & \textbf{48.94}  & 23.91            & 39.74            \\ \cline{3-9} 
                           &                             & Adjacent Instruction & 70.00            & 61.57             & 35.87           & 42.55           & 64.13            & 54.82            \\
                           
                           &                             & + Sup-Scorer (Max)   & 66.00            & 45.52             & 30.43           & 42.55           & 51.09            & 47.12            \\
                           &                             & + Sup-Scorer (Avg)   & 66.00            & 61.94             & 31.52           & 39.36           & 71.74            & 54.11            \\
                           &                             & + Adjacent Answer      & 69.00            & \textbf{65.67}    & \textbf{39.13}  & 39.36           & \textbf{72.83}   & \textbf{57.20}   \\
                           \cline{3-9} 
                           \hline
\end{tabular}}
\caption{Experiment results of different negative sampling methods for contrastive training.}
\label{tab:negative-sampling}
\end{table}

As shown in Table \ref{tab:negative-sampling}, the superficial scorer does not introduce improvement. This notifies us that the superficial quality can not be simply attributed solely to the target side. While target-only classifier being verified as an effective bias measurement for traditional translation evaluation, the situation for LLM evaluation is more complicated.

On the other hand, the adjacent selection based both on the instruction and the answer achieves better results. This suggests the instruction similarity modeling and answer similarity modeling obtain information of the original data from different perspectives, which promotes the training of the judge model. The instruction similarity can help understand the nuances of the tasks, while the answer similarity can help understand the solution approaches. Therefore, it is better to perform negative sampling both on the instructions and answers.

\section{Conclusion}
In this work, we propose two methods for mitigating the bias of LLM-as-a-Judge. For closed-source judge models, we propose to mitigate the bias by calibration. For open-source judge models, we propose to mitigate the bias by pairwise contrastive training. We apply our methods on the LLMBar benchmark, effectively mitigating the bias of LLM evaluation while maintaining a comparable performance.

As the LLM-as-a-Judge becomes a paradigm for evaluating the LLM performance, it is advisable to exert more alert on the bias of LLM-based evaluators. In the future, we will dedicate more effort to the bias mitigation of LLM-as-a-Judge into more fine-grained categories. We will also conduct research about the inference bias on other LLM-based applications.

\section*{Acknowledgements}

This work is supported by National Natural Science Foundation of China (62376075, 62276077, U1908216, 62376076), Key R\&D Program of Yunnan (202203AA080004) and Shenzhen College Stability Support Plan (No. GXWD20220811170358002).

\bibliographystyle{ccl}
\bibliography{custom}

\end{document}